# Freedom: A Measure of Second-Order Uncertainty for Intervalic Probability Schemes

Dr. Michael Smithson, Behavioural Sciences, James Cook University
Queensland 4811   Australia

## Abstract

This paper discusses a new measure that is adaptable to certain intervalic probability frameworks, possibility theory, and belief theory. As such, it has the potential for wide use in knowledge engineering, expert systems, and related problems in the human sciences. This measure (denoted here by F) has been introduced in Smithson (1988) and is more formally discussed in Smithson (1989a). Here, I propose to outline the conceptual basis for F and compare its properties with other measures of second-order uncertainty. I will argue that F is an indicator of *nonspecificity* or alternatively, of *freedom*, as distinguished from either ambiguity or vagueness.

## 1. A Measure of Nonspecificity and Freedom

Recently, significant advances have been made in the formal representation of second-order uncertainty (starting with Shafer's (1976) belief theory and Zadeh's (1978) possibility theory). These developments followed on from earlier work in second-order probability (e.g., C.A.B. Smith 1961), and there have been attempts to unify the various approaches (e.g., Loui 1986, Dubois and Prade 1987, and Kyburg 1987). These new formalisms have brought into focus a new set of problems concerning the definition and measurement of various kinds of second-order uncertainty. While various terms and measures have been proposed for second-order uncertainty, as yet the area lacks a widely accepted unifying framework. Moreover, it is not clear that all of the important kinds of second-order uncertainty have been defined and operationalized. In particular, nonspecificity has not been distinguished from vagueness or ambiguity (and the use of these terms in the literature is unsatisfactory). This paper presents a definition and measure of nonspecificity, denoted here by F.

Without loss of generality, I shall introduce F in the setting of possibility theory. Accordingly, let p(E) be the probability of E, and let $ne(E) \leq p(E) \leq po(E)$, where ne denotes necessity and po denotes possibility. In the absence of any further information about p(E) and its range of permissible values, we assume that p(E) has a uniform distribution over the interval [ne(E),po(E)]. A unimodal distribution of p(E) would be more *specific* than this condition, while a multi-modal distribution would tend towards *ambiguity*.

An illustration should suffice to make this last point. Gardenfors and Sahlins (1982) asks us to consider Miss Julie who has been invited to bet on the outcomes of three tennis matches. In Match A, she is informed that the players are very close in ability and so she has good reason to believe that the probability of either player winning is close to 0.5. She is uninformed, however, about the players in Match B and so cannot say anything about the probability that either of them will win. Prior to Match C, she overhears someone remark that one of the players is extremely good but the other is a novice. Unfortunately, she does not know which player is the better one, so she cannot tell which way to bet.

These are three quite distinct second-order probability distributions. Match A connotes a tight interval around 0.5. Match B indicates a uniform distribution spanning the [0,1] interval. Match C corresponds to a sharply bimodal distribution, with the distribution concentrated near 0 and 1. Clearly the distribution of probabilities in Match B is the most *nonspecific* of the three (in the sense of Max Black's 1937 paper), while Match A has a quite specific distribution. The distribution in Match C also is more specific than that for Match B, and corresponds to the usual linguistic or philosophical definition of *ambiguity*. Therefore, Klir's (1987) version of 'ambiguity' is too inclusive; he fails to distinguish it from nonspecificity. We should not expect that the measures of 'ambiguity' reviewed in his survey would adequately measure nonspecificity (and indeed they do not).

Under conditions where the second-order distribution of p(E) is not specified beyond its upper and lower bounds, or in ordinary possibility theory, it is reasonable to define a measure of *relative nonspecificity* by

$$F = po(E) - ne(E) \tag{1}$$

which corresponds to Baldwin's (1986) so-called measure of 'uncertainty' in his support logic system. In Smithson (1988) I provide a frequentist rationale for F as a measure of the relative *freedom* of a collection of elements to 'choose' E or not. If, say, 15% of a collection of people must use their automobiles to get to work, and 75% of them could possibly do so, then the percentage who are 'free' to choose whether or not they will use their automobiles to go to work is 75%-15% = 60%.

More generally, consider M options, denoted by $E_i$, $1 \leq i \leq M$. The definition of F must take into account the restriction that the $p_i$ must sum to 1. A geometric representation of this system is a M-1 dimensional simplex whose edges are defined by the sum-constraint. The $po_i$ and $ne_i$ remove 'slices' of this simplex, and F may therefore be defined as the proportion of the volume remaining (a proof is given in Smithson 1989a):

$$F_M = (1 - \sum_{i=1}^{M} ne_i)^{M-1} - \sum_{i=1}^{M} \max(0, 1 - po_i - \sum_{j \neq i} ne_j)^{M-1}$$

$$+ \sum_{i=1}^{M} \sum_{j>i} \max(0, 1-po_i-po_j-\sum_{k\neq i,j} ne_k)^{M-1} - \ldots$$

$$\ldots + (-1)^{M-1} \sum_{i=1}^{M} \max(0, 1-ne_i-\sum_{j\neq i} po_j)^{M-1} \tag{2}$$

For example, for $M = 3$, we have
$$F_3 = (1-ne_1-ne_2-ne_3)^2 - \max(0,1-po_1-ne_2-ne_3)^2$$
$$- \max(0,1-po_2-ne_1-ne_3)^2 - \max(0,1-po_3-ne_1-ne_2)^2$$
$$+ \max(0, 1-po_1-po_2-ne_3)^2 + \max(0,1-po_1-po_3-ne_2)^2$$
$$+ \max(0,1-po_2-po_3-ne_1)^2 \tag{3}$$

Figure 1 displays a geometric representation of $F_3$ as the area left in the triangle defined by $p_1+p_2 \leq 1$ when the $po_i$ and $ne_i$ are taken into account. Clearly $F_M$ attains 1 iff for all i $po_i = 1$ and $ne_i = 0$, which corresponds to total nonspecificity (i.e., total ignorance from a subjectivist standpoint, and total freedom from a frequentist perspective). Likewise, it attains 0 iff for any i, $po_i = ne_i$, or in other words when any of the $p_i$ are totally specified.

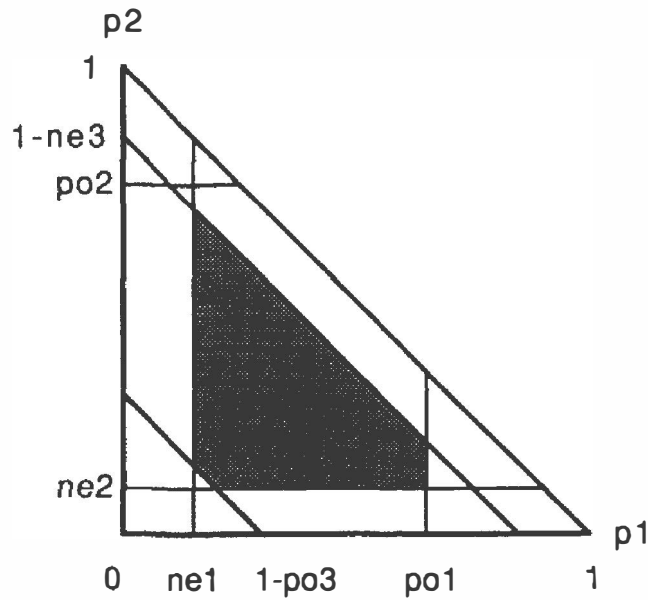

**Figure 1. Three-Dimensional Example**

Let us compare $F_M$ with measures of so-called 'ambiguity' which have been proposed in the recent literature. One is due to Yager (1982), and may be written in a form following Klir (1987):

$$A_M = 1 - \sum_{i=1}^{M} \frac{po_i - po_{i+1}}{i} \tag{4}$$

where $po_i > po_{i+1}$ for all i, and by convention $po_{n+1} = 0$. The second is a generalization of Hartley's information theoretic measure (Higashi and Klir 1983):

$$I_M = \sum_{i=1}^{M} (po_i - po_{i+1}) \log_2 i \tag{5}$$

While $I_M$ and $A_M$ are clearly monotonically related, $F_M$ is not monotonically related to either of them.

Example 1. For M=2, consider two cases:
  Case 1. $po_1 = 0.8$, $po_2 = 0.4$ (whence $ne_1 = 0.6$, $ne_2 = 0.2$)
  Case 2. $po_1 = 1.0$, $po_2 = 0.4$ (whence $ne_1 = 0.6$, $ne_2 = 0.0$)
In Case 1, $A_2 = 0.4$, while in Case 2, $A_2 = 0.2$.
In Case 1, $F_2 = 0.2$, while in Case 2, $F_2 = 0.4$.

Moreover, in terms of the 'latitude' represented in these two cases, $F_2$ yields results that are consistent with intuition while $A_2$'s values run counter to intuition.

Example 2. For M = 2, consider two cases:
  Case 1. $po_1 = 1.0$, $po_2 = 0.6$ (whence $ne_1 = 0.4$, $ne_2 = 0.0$)
  Case 2. $po_1 = 0.7$, $po_2 = 0.7$ (whence $ne_1 = 0.3$, $ne_2 = 0.3$)
In Case 1, $I_2 = 0.6$, while in Case 2, $I_2 = 0.7$.
In Case 1, $F_2 = 0.6$, while in Case 2, $F_2 = 0.4$.

Again $F_2$ yields values in line with intuition, while $I_2$ does not.

$F_M$ also increases whenever any $po_i$ is increased, provided that $\max_j(ne_j) < 1-po_i$ for j not equal to i, or if at least one of the $ne_j$ is decremented accordingly (which it has to be in order to maintain a consistent assignment of $po_i$'s and $ne_i$'s). Otherwise, $F_M$ stays constant. However, for any M, $I_M$ stays constant when $po_1$ increases (it ignores $po_1$ by virtue of the fact that $\log(1) = 0$). This is quite counterproductive if we wish to depend on $I_M$ as a measure of nonspecificity or freedom. For instance, it fails to distinguish the case $\{po_1 = po_2 = po_3 = 0.5\}$ from $\{po_1 = 1.0, po_2 = po_3 = 0.5\}$. In both instances $I_3 = 0.5\log_2(3)$, whereas $F_3 = 0.25$ for the first case and 0.50 in the second (thereby differing by a factor of 2).

It can be shown that $F_M$ is maximal only under total ignorance, and minimal for probability measures in any subset of the M options, whereas $A_M$ and $I_M$ attain their minima only for the case where there is only one i such that $po_i = 1$ and all other $poj = 0$. Clearly $A_M$ and $I_M$ are nonzero even when $po_i = ne_i$ for all i, as long as there is more than one i for which there are nonzero values. In that respect, $A_M$ and $I_M$ may fail to distinguish mere first-order uncertainty from second-order uncertainty. $F_M$, on the other hand, always attains 0 when there is no second-order uncertainty.

One implication of the condition under which $F_M$ attains 0 is that for M>2, one needs to consider not only $F_M$ but also $F_K$, where K runs over all subsets of M. Moreover, it may be useful in some applications to consider a conditional version of $F_K$, in which the condition is that some proportion (p, say) of belief has been assigned in the form of pointwise (rather than intervalic) probabilities for some subset of M-K alternatives. Under that condition, a conditional version of $F_K$ may be defined by computing the portion of the volume of the right-simplex whose height is $q = 1-p$. Denoting this conditional measure by $F_{K|q}$, we define it by replacing all occurrences of 1 in the major terms of (2) with q.

Finally, we must be wary of comparisons between $F_M$ and $F_N$, where M and N are unequal. As pointed out in Smithson (1989a), there is a connection between $F_M$ and the traditional statistical concept of 'degrees of freedom', in that $F_M$ measures the fraction of the total freedom that could be realized given M degrees of freedom as a maximum potential. Thus, the meaning of the magnitude of $F_M$ is conditional on M itself. A partial remedy for this incommensurability is to use $S_M = (F_M)^{1/(M-1)}$ as a 'normed' measure, although even then we are still dealing with a measure of nonspecificity (or freedom) given M options.

## 2. Effects of $po_i$ and $ne_i$ on $F_M$

Whether $F_M$ is viewed as a utility (as in freedom) or a disutility (as in nonspecificity), it is reasonable to ask whether an change in the $po_i$ has the same effect on $F_M$ as an equal and opposite change in the $ne_i$. Given the choice, when should we make an effort to alter the $po_i$ and when should we concentrate on the $ne_i$?

It can be shown (Smithson 1989a, Theorem 5) that changes in the $ne_i$ have a greater impact on $F_M$ than equal and opposite changes in the $po_i$ unless the following condition holds: For some subset of the M options whose cardinality is M-1, $\sum_{i=1}^{M-1} ne_i < 1 - \sum_{i=1}^{M-1} po_i$. In that event, a change in $po_M$ will have a greater impact on FM than an equal and opposite change in $ne_M$. Figure 2 shows an example for M=3. The left-hand part corresponds to a case where the above condition holds, and the right-hand part to a case where it does not. The shaded areas indicate the amount of $F_M$ lost by imposing $po_3$ and $ne_3$, where $1-po_3 = ne_3$. Clearly, in the left-hand instance $po_3$ subtracts the greater area, while in the right-hand instance $ne_3$ subtracts more.

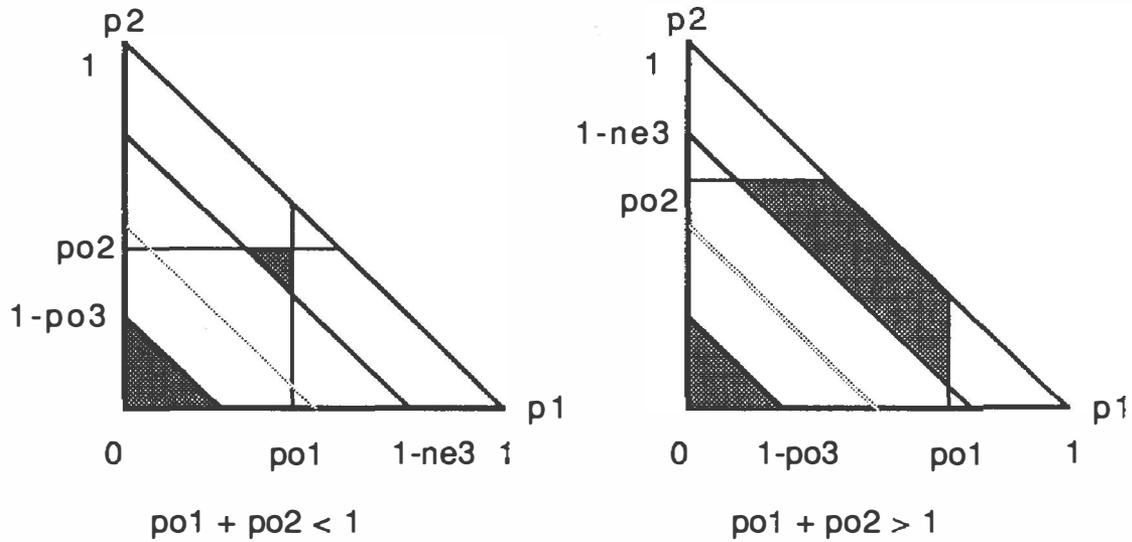

Figure 2. Effects of $ne_3$ and $po_3$ on $F_3$

It should be evident that for small M, the conditions specified above may not be uncommon, while for larger M it will tend to be rarer. In fact, in general this condition requires rather 'low' values for the $po_i$. The reason is that it implies the sum of the M-1 $po_i$ must be less than 1, which means that the requirement that the sum of all M $po_i$ be greater than 1 would be achieved only by adding in the last, $po_M$.

### 3. Nonspecificity in Cross-Classified Systems

In some kinds of applications we may wish to consider cross-classified systems. These are a special case of systems in which there are po and ne constraints on sums across subsets of the $p_i$. Consider a K-option system of $E_i$ which is cross-classified with an M-option system of $E_j$. Let $p_{ij}$ denote the joint probability of $E_i$ and $E_j$, and $p_{i.}$ and $p_{.j}$ the respective marginal probabilities. Then, as Baldwin (1986) observes, given $ne_{i.} \leq p_{i.} \leq po_{i.}$ and $ne_{.j} \leq p_{.j} \leq po_{.j}$, we have

$ne(ne_{ij}) = \max(0, ne_{i.}+ne_{.j}-1) \leq ne_{ij} \leq \min(ne_{i.},ne_{.j}) = po(ne_{ij})$, and

$ne(po_{ij}) = \max(0, po_{i.}+po_{.j}-1) \leq po_{ij} \leq \min(po_{i.},po_{.j}) = po(po_{ij})$.  (6)

Clearly it is of interest to evaluate not only the marginal nonspecificity measures $F_K$ and $F_M$, but also the joint nonspecificity $F_{K \times M}$. The latter is not straightforward, and we do not have an explicit expression for the general case. However, there are practical algorithms for evaluating these measures (e.g., Cohen and Hickey 1979).

Likewise, $F_{K \times M}$ and, for that matter, the nonspecificity of individual $p_{ij}$, depend partly on the nature of the (in)dependence between the $E_i$ and $E_j$. Let $D_{ij}$ denote the degree to which $E_i$ and $E_j$ are dependent, and define it by $D_{ij} = (p_{ij}-B_{ij})/(A_{ij}-B_{ij})$, where $A_{ij} = \min(p_{i.},p_{.j})$ and $B_{ij} = \max(0, p_{i.}+p_{.j}-$ 1). Thus, $D_{ij} = 1$ when $E_i$ and $E_j$ are maximally overlapped and $D_{ij} = 0$ when they are maximally disjoint (note that $D_{ij} = 1/2$ does not mean statistical independence, however). It can be shown (cf. Smithson 1989a) that the nonspecificity of $p_{ij}$ is maximized or minimized whenever $D_{ij} = 0$ or 1, according to the following cases:

Case 1: If $ne_{i.} + ne_{.j} > 1$, then nonspecificity is maximized when $D_{ij} = 0$ and minimized when $D_{ij} = 1$. Moreover, this can occur only for *one* cell in any cross-classification of any size. For suppose $ne_{i.} + ne_{.j} > 1$. Then $\sum_{k \neq i} ne_{k.} \leq 1 - ne_{i.}$ and $\sum_{m \neq j} ne_{.j} \leq 1 - ne_{.j}$, so $\sum_{k \neq i} ne_{k.} + \sum_{m \neq j} ne_{.j} \leq 2 - (ne_{i.} + ne_{.j}) < 1$.

Case 2: If $po_{i.} + po_{.j} < 1$, then nonspecificity is maximized when $D_{ij} = 1$ and minimized when $D_{ij} = 0$. This condition may occur for all cells if M and K both are > 2. Otherwise, there must be at least one cell in the table for which Case 2 does not hold. Moreover, we should expect this case to become more common as M and K become large.

Case 3: Suppose $ne_{i.} + ne_{.j} \leq 1$ and $po_{i.} + po_{.j} \geq 1$. Then if $1 - \min(ne_{i.}, ne_{.j}) >(<) \max(po_{i.}, po_{.j})$, nonspecificity is maximized (minimized) when $D_{ij} = 0$ and minimized (maximized) when $D_{ij} = 1$. In larger tables, for reasonably evenly distributed $p_{ij}$, one would expect that most of the time $1 - \min(ne_{i.}, ne_{.j}) > \max(po_{i.}, po_{.j})$.

These cases demonstrate that the nonspecificity of cross-classified systems varies as a function of their marginal nonspecificity and the dependency between the $E_i$ and $E_j$. This result has several implications for the evaluation of uncertainty in complex systems. First, the proliferation of this kind of uncertainty through time in a given system will depend on how tightly coupled its various stages are, as well as how unspecified the marginals are for each of those stages. Secondly, in systems where the dependency relations are not known, any sensitivity analysis of nonspecificity must include variations in dependency.

## 4. Second-Order Uncertainty and Human Judgment

The foregoing discussion has introduced a new measure of second-order uncertainty with an appealing rationale, distinct from second-order ambiguity and fuzziness. The measure has potential for many applications, and practical computing algorithms are available even for intractable cases. In closing, I should like to very briefly indicate a few implications and findings stemming from recent empirical work on F and its associated concepts of freedom and nonspecificity.

First, a recent study (Smithson 1989b) finds that people are averse to second-order vagueness, nonspecificity, and ambiguity in the probabilities associated with gambles involving the prospect of gain. Therefore, we have a second-order version of Ellsberg's Paradox. Secondly, current experiments suggest that, given a constant $F_M$, a system's performance is evaluated more positively when nonspecificity is represented in terms of desirable outcomes (e.g., successes) than when it is represented in terms of undesirable outcomes (e.g., failures). Another indicates that the degree of nonspecificity itself is perceived as greater when represented in terms of possibility than when represented in terms of necessity (constraint). These results, although tentative, highlight the importance of furthering our understanding of human judgment and decision making when freedom or nonspecificity is a consideration, as well as the usefulness of a normatively defensible measure of nonspecificity.

## References

Baldwin, J.F. 1986 "Support logic programming." In A.I. Jones, et al. (eds.) *Fuzzy Sets: Theory and Applications*. Dordrecht: Reidel.

Black, M. 1937 "Vagueness: An exercise in logical analysis." *Philosophy of Science*, 4: 427-455.

Cohen, J. and Hickey, T. 1979 "Two algorithms for determining volumes of convex polyhedra." *Journal of the Association for Computing Machinery*, 26: 401-414.

Dubois, D. and Prade, H. 1987b "Properties of measures of information in evidence and possibility theories." *Fuzzy Sets and Systems*, 24: 161-182.

Gardenfors, P. and Sahlins, N.-E. 1982 "Unreliable probabilities, risk taking, and decision making." *Synthese*, 53: 361-386.

Higashi, M. and Klir, G.J. 1983 "On the notion of distance representing information closeness: possibility and probability distributions." *International Journal of General Systems*, 9: 103-115.

Klir, G.J. 1987 "Where do we stand on measures of uncertainty, ambiguity, fuzziness, and the like?" *Fuzzy Sets and Systems*, 24: 141-160.

Kyburg, H. 1987 "Representing knowledge and evidence for decision." In B. Bouchon and R.R. Yager (eds.) *Uncertainty in Knowledge-Based Systems: Lecture Notes in Computer Science No. 286*. New York: Springer Verlag.

Loui, R.P. 1986 "Interval-based decisions for reasoning systems." In L.N. Kanal and J.F. Lemmer (eds.) *Uncertainty in Artificial Intelligence*. Amsterdam: North-Holland.

Shafer, G. 1976 *A Mathematical Theory of Evidence*. Princeton: Princeton University Press.

Smith, C.A.B. 1961 "Consistency in statistical inference and decision." *Journal of Royal Statistical Society, Series B*, 23:1-37.

Smithson, M. 1988 "Possibility theory, fuzzy logic, and psychological explanation." In T. Zetenyi (ed.) *Fuzzy Sets in Psychology*. Amsterdam: North-Holland.

Smithson, M. 1989a "Measures of freedom based on possibility." *Mathematical Social Sciences*, 18: in press.

Smithson, M. 1989b *Ignorance and Uncertainty: Emerging Paradigms*. New York: Springer Verlag, 393 pp.

Yager, R.R. 1982 "Measuring tranquility and anxiety in decision making: An application of fuzzy sets." *International Journal of General Systems*, 8: 139-146.

Zadeh, L.A. 1978 "Fuzzy sets as a basis for a theory of possibility." *Fuzzy Sets and Systems*, 1: 3-28.